\documentclass{article} % For LaTeX2e

\usepackage{nips14submit_e,times}
\usepackage{epsf}
\usepackage{tabulary}
\usepackage{subfigure}
\usepackage{tabularx}
\usepackage{amssymb}
\usepackage{subfigure}
\usepackage{graphicx}
\usepackage{wrapfig}
\usepackage{amsmath}
\usepackage{url}
\usepackage{algorithm}
\usepackage{algorithmic}
\usepackage{enumitem}
\usepackage{epstopdf}
\usepackage[T1]{fontenc}
\usepackage{bm}
\usepackage{eqparbox}
\usepackage[numbers]{natbib}

\usepackage{graphicx} % more modern
\usepackage{subfigure} 

\usepackage{natbib}

\usepackage{algorithm}
\usepackage{algorithmic}

\title{Unifying Visual-Semantic Embeddings with Multimodal Neural Language Models}

\author{
Ryan Kiros, Ruslan Salakhutdinov, Richard S. Zemel \\
University of Toronto \\
Canadian Institute for Advanced Research \\
\texttt{\{rkiros, rsalakhu, zemel\}@cs.toronto.edu} \\
}

\nipsfinalcopy % Uncomment for camera-ready version

\begin{document}

\maketitle

\begin{abstract}
Inspired by recent advances in multimodal learning and machine translation, we introduce an encoder-decoder pipeline that
learns (a): a multimodal joint embedding space with images and text and (b): a novel language model for decoding
distributed representations from our space. Our pipeline effectively unifies joint image-text embedding models
with multimodal neural language models. We introduce the structure-content neural language model that disentangles
the structure of a sentence to its content, conditioned on representations produced by the encoder. The encoder allows one to rank
images and sentences while the decoder can generate novel descriptions from scratch.  Using LSTM to encode sentences, we match the state-of-the-art performance on Flickr8K and Flickr30K without using object detections. We also set new best results when using the 19-layer Oxford convolutional network. Furthermore we show that with linear encoders, the learned embedding space captures 
multimodal regularities in terms of vector space arithmetic e.g. *image of a blue car* - "blue" + "red" is near 
images of red cars. Sample captions generated for 800 images are made available for comparison.
\end{abstract}

\section{Introduction}
\label{intro}

Generating descriptions for images has long been regarded as a challenging perception task integrating vision, learning and language understanding. One not only needs to correctly recognize what appears in images but also incorporate knowledge of spatial relationships and interactions between objects. Even with this information, one then needs to generate a description that is relevant and grammatically correct. With the recent advances made in deep neural networks, tasks such as object recognition and detection have made significant breakthroughs in only a short time. The task of describing images is one that now appears tractable and ripe for advancement. Being able to append large image databases with accurate descriptions for each image would significantly improve the capabilities of content-based image retrieval systems. Moreover, systems that can describe images well, could in principle, be fine-tuned to answer questions about images also.

This paper describes a new approach to the problem of image caption generation, casted into the framework of encoder-decoder models. For the encoder, we learn a joint image-sentence embedding where sentences are encoded using long short-term memory (LSTM) recurrent neural networks \cite{hochreiter1997long}. Image features from a deep convolutional network are projected into the embedding space of the LSTM hidden states. A pairwise ranking loss is minimized in order to learn to rank images and their descriptions. For decoding, we introduce a new neural language model called the structure-content neural language model (SC-NLM). The SC-NLM differs from existing models in that it disentangles the structure of a sentence to its content, conditioned on distributed representations produced by the encoder. We show that sampling from an SC-NLM allows us to generate realistic image captions, significantly improving over the generated captions produced by \cite{kirosmultimodal}. Furthermore, we argue that this combination of approaches naturally fits into the experimentation framework of \cite{hodosh2013framing}, that is, a good encoder can be used to \textit{rank} images and captions while a good decoder can be used to \textit{generate} new captions from scratch. Our approach effectively unifies image-text embedding models (encoder phase) \cite{weston2010large, fromedevise2013, socher2013grounded} with multimodal neural language models (decoder phase) \cite{kirosmultimodal} \cite{mao2014explain}. Furthermore, our method builds on analogous approaches being used in machine translation \cite{kalchbrenner2013recurrent, cho2014learning, bahdanau2014neural, sutskever2014sequence}.

While the application focus of our work is on image description generation and ranking, we also qualitatively analyse properties of multimodal vector spaces learned using images and sentences. We show that using a linear sentence encoder, linguistic regularities \cite{mikolov2013linguistic} also carry over to multimodal vector spaces. For example, *image of a blue car* - "blue" + "red" results in a vector that is near images of red cars. We qualitatively examine several types of analogies and structures with PCA projections. Consequently, even with a global image-sentence training objective the encoder can still be used to retrieve locally (e.g. individual words). This is analogous to pairwise ranking methods used in machine translation \cite{hermann2013multilingual, Hermann:2014:ACLphil}.

\begin{figure}
\centering
\includegraphics[width=5.0in]{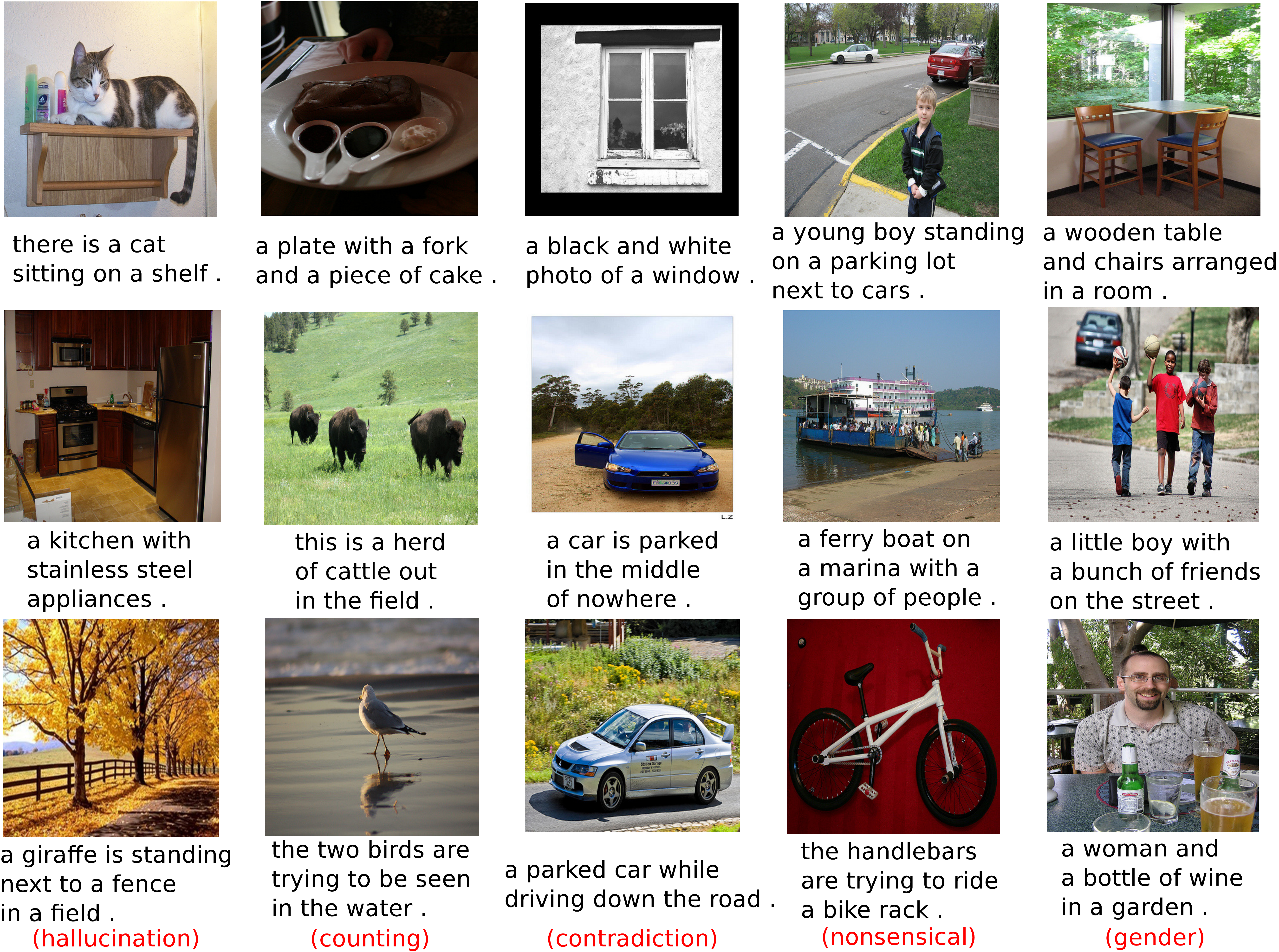} 
\caption{Sample generated captions. The bottom row shows different error cases. Additional results can be found at \url{http://www.cs.toronto.edu/\textasciitilde rkiros/lstm_scnlm.html}}
\label{fig:annot}
\end{figure}

\subsection{Multimodal representation learning}

A large body of work has been done on learning multimodal representations of images and text. Popular approaches include learning joint image-word embeddings \cite{weston2010large, fromedevise2013} as well as embedding images and sentences into a common space \cite{socher2013grounded, karpathy2014deep}. Our proposed pipeline makes direct use of these ideas. Other approaches to multimodal learning include the use of deep Boltzmann machines \cite{srivastava2012multimodal}, log-bilinear neural language models \cite{kirosmultimodal}, autoencoders \cite{ngiam2011multimodal}, recurrent neural networks \cite{mao2014explain} and topic-models \cite{jia2011learning}. Several bi-directional approaches to ranking images and captions have also been proposed, based off of kernel CCA \cite{hodosh2013framing}, normalized CCA \cite{gong2014improving} and dependency tree recursive networks \cite{socher2013grounded}. From an architectural standpoint, our encoder-decoder model is most similar to \cite{blunsom2014deep}, who proposed a two-step embedding and generation procedure for semantic parsing. 

\subsection{Generating descriptions of images}

We group together approaches to generation into three types of methods, each described here in more detail:

\textbf{Template-based methods.} Template-based methods involve filling in sentence templates, such as triplets, based on the results of object detections and spatial relationships \cite{kulkarni2011baby, farhadi2010every, li2011composing, yang2011corpus, mitchell2012midge}. While these approaches can produce accurate descriptions, they are often more `robotic' in nature and do not generalize to the fluidity and naturalness of captions written by humans.

\textbf{Composition-based methods.} These approaches aim to harness existing image-caption databases by extracting components of related captions and composing them together to generate novel descriptions \cite{kuznetsova2012collective, berg2014tree}. The advantage of these approaches are that they allow for a much broader and more expressive class of captions that are more fluent and human-like then template-based approaches. 

\textbf{Neural network methods.} These approaches aim to generate descriptions by sampling from conditional neural language models. The initial work in this area, based off of multimodal neural language models \cite{kirosmultimodal}, generated captions by conditioning on feature vectors from the output of a deep convolutional network. These ideas were recently extended to multimodal recurrent networks with significant improvements \cite{mao2014explain}. The methods described in this paper produce descriptions that at least qualitatively on par with current state-of-the-art composition-based methods \cite{berg2014tree}.

Description generation systems have been plagued with issues of evaluation. While Bleu and Rouge have been used in the past, \cite{hodosh2013framing} has argued that such automated evaluation methods are unreliable and do not match human judgements. These authors instead proposed that the problem of ranking images and captions can be used as a proxy for generation. Since any generation system requires a scoring function to access how well a caption and image match, optimizing this task should naturally carry over to an improvement in generation. Many recent methods have since used this approach for evaluation. None the less, the question on how to transfer improvements on ranking to generating new descriptions remained. We argue that encoder-decoder methods naturally fit into this experimentation framework. That is, the encoder gives us a way to rank images and captions and develop good scoring functions, while the decoder can use the representations learned to optimize the scoring functions as a way of generating and scoring new descriptions.

\subsection{Encoder-decoder methods for machine translation}

\begin{figure}
\centering
\includegraphics[width=5in]{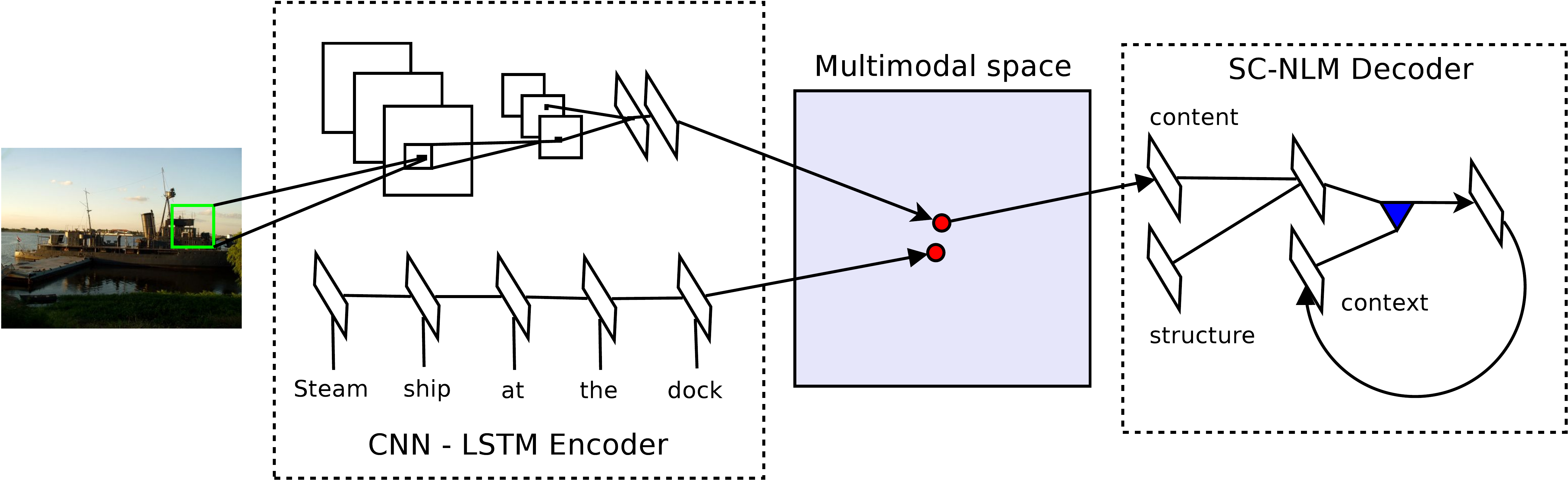} 
\caption{\textbf{Encoder:} A deep convolutional network (CNN) and long short-term memory recurrent network (LSTM) for learning a joint image-sentence embedding. \textbf{Decoder:} A new neural language model that combines structure and content vectors for generating words one at a time in sequence.}
\label{fig:annot}
\end{figure}

Our proposed pipeline, while new to caption generation, has already experienced several successes in Neural Machine Translation (NMT). The goal of NMT is to develop an end-to-end translation system with a large neural network, as opposed to using a neural network as an additional feature function to an existing phrase-based system. NMT methods are based on the encoder-decoder principle. That is, an encoder is used to map an English sentence to a distributed vector. A decoder is then conditioned on this vector to generate a French translation from the source text. Current methods include using a convolutional encoder and RNN decoder \cite{kalchbrenner2013recurrent}, RNN encoder and RNN decoder \cite{cho2014learning, bahdanau2014neural} and LSTM encoder with LSTM decoder \cite{sutskever2014sequence}. While still a young research area, these methods have already achieved performance on par with strong phrase-based systems and have improved on the start-of-the-art when used for rescoring.

We argue that it is natural to think of image caption generation as a translation problem. That is, our goal is to \textit{translate} an image into a description. This point of view has also been used by \cite{rohrbach2013translating} and allows us to make use of existing ideas in the machine translation literature. Furthermore, there is a natural correspondence between the concept of scoring functions (how well does a caption and image match) and alignments (which parts of a description correspond to which parts of an image) that can naturally be exploited for generating descriptions.

\section{An encoder-decoder model for ranking and generation}

In this section we describe our image caption generation pipeline. We first review LSTM RNNs which are used for encoding sentences, followed by how to learn multimodal distributed representations. We then review log-bilinear neural language models \cite{mnih2007three}, multiplicative neural language models \cite{kiros2014multiplicative} and then introduce our structure-content neural language model.

\subsection{Long short-term memory RNNs}

Long short-term memory \cite{hochreiter1997long} is a recurrent neural network that incorporates a built in memory cell to store information and exploit long range context. LSTM memory cells are surrounded by gating units for the purpose of reading, writing and reseting information. LSTMs have been used to achieve state-of-the-art performance in several tasks such as handwriting recognition \cite{graves2009novel}, sequence generation \cite{graves2013generating} speech recognition \cite{graves2013hybrid} and machine translation \cite{sutskever2014sequence} among others. Dropout \cite{srivastava2014dropout} strategies have also been proposed to prevent overfitting in deep LSTMs. \cite{zaremba2014recurrent}

Let ${\bf X}_t$ denote a matrix of training instances at time $t$. In our case, ${\bf X}_t$ is used to denote a matrix of word representations for the $t$-th word of each sentence in the training batch. Let $({\bf I}_t, {\bf F}_t, {\bf C}_t, {\bf O}_t, {\bf M}_t)$ denote the input, forget, cell, output and hidden states of the LSTM at time step $t$. The LSTM architecture in this work is implemented using the following equations:
\begin{eqnarray}
{\bf I}_t &=& \sigma({\bf X}_t \cdot {\bf W}_{xi} + {\bf M}_{t-1} \cdot {\bf W}_{hi} + {\bf C}_{t-1} \cdot {\bf W}_{ci} + {\bf b}_i) \\
{\bf F}_t &=& \sigma({\bf X}_t \cdot {\bf W}_{xf} + {\bf M}_{t-1} \cdot {\bf W}_{hf} + {\bf C}_{t-1} \cdot {\bf W}_{cf} + {\bf b}_f) \\
{\bf C}_t &=& {\bf F}_t \bullet {\bf C}_{t-1} + {\bf I}_t \bullet tanh( {\bf X}_t \cdot {\bf W}_{xc} + {\bf M}_{t-1} \cdot {\bf W}_{hc} + {\bf b}_c) \\
{\bf O}_t &=& \sigma({\bf X}_t \cdot {\bf W}_{xo} + {\bf M}_{t-1} \cdot {\bf W}_{ho} + {\bf C}_t \cdot {\bf W}_{co} + {\bf b}_o) \\
{\bf M}_t &=& {\bf O}_t \bullet tanh({\bf C}_t)
\end{eqnarray}
where ($\sigma$) denotes the sigmoid activation function, ($\cdot$) indicates matrix multiplication and ($\bullet$) indicates component-wise multiplication. \footnote{For additional details on LSTM: \url{http://people.idsia.ch/~juergen/rnn.html}.}

\subsection{Multimodal distributed representations}

Suppose for training we are given image-description pairs each corresponding to an image and a description that correctly describes the image. Images are represented as the top layer (before the softmax) of a convolutional network trained on the ImageNet classification task \cite{krizhevsky2012imagenet}. 

Let $D$ be the dimensionality of an image feature vector (e.g. 4096 for AlexNet \cite{krizhevsky2012imagenet}), $K$ the dimensionality of the embedding space and let $V$ be the number of words in the vocabulary. Let ${\bf W}_I \in \mathbb{R}^{K \times D}$ and ${\bf W}_T \in \mathbb{R}^{K \times V}$ be the image embedding matrix and word embedding matrices, respectively. Given an image description $S = \{w_1, \ldots, w_N\}$ with words $w_1, \ldots, w_N$, \footnote{As a slight abuse of notation, we refer to $w_i$ as both a word and an index into the word embedding matrix.}
let $\{ {\bf w}_1, \ldots, {\bf w}_N \}, {\bf w}_i \in \mathbb{R}^K, i=1,\ldots,n$ denote the corresponding word representations to words $w_1, \ldots, w_N$ (entries in the matrix ${\bf W}_T$). The representation of a sentence ${\bf v}$ is the hidden state of the LSTM at time step $N$ (i.e. the vector ${\bf m}_t$). We note that other approaches for computing sentence representations for image-text embeddings have been proposed, including dependency tree RNNs \cite{socher2013grounded} and bags of dependency parses \cite{karpathy2014deep}. Let ${\bf q} \in \mathbb{R}^D$ denote an image feature vector (for the image corresponding to description $S$) and let ${\bf x} = {\bf W}_I \cdot {\bf q} \in \mathbb{R}^K$ be the image embedding. We define a scoring function $s({\bf x}, {\bf v}) = {\bf x} \cdot {\bf v}$, where ${\bf x}$ and ${\bf v}$ are first scaled to have unit norm (making $s$ equivalent to cosine similarity). Let $\boldsymbol\theta$ denote all the parameters to be learned (${\bf W}_I$ and all the LSTM weights) \footnote{We keep the word embedding matrix ${\bf W}_T$ fixed.}. We optimize the following pairwise ranking loss:
\begin{eqnarray}
\underset{\boldsymbol\theta}{\operatorname{min}} \hspace{1mm}
\sum_{\bf x} \sum_k \text{max}\{0, \alpha - s({\bf x}, {\bf v}) + s({\bf x}, {\bf v}_k)\} + 
\sum_{\bf v} \sum_k \text{max}\{0, \alpha - s({\bf v}, {\bf x}) + s({\bf v}, {\bf x}_k)\} 
\end{eqnarray}
where ${\bf v}_k$ is a contrastive (non-descriptive) sentence for image embedding ${\bf x}$, and vice-versa with ${\bf x}_k$. For all of our experiments, we initialize the word embeddings ${\bf W}_T$ to be pre-computed $K = 300$ dimensional vectors learned using a continuous bag-of-words model \cite{mikolov2013efficient}. The contrastive terms are chosen randomly from the training set and resampled every epoch.

\subsection{Log-bilinear neural language models}

The log-bilinear language model (LBL) \citep{mnih2007three} is a 
deterministic model that may be viewed as a feed-forward neural network with a single linear hidden layer. Each word $w$ in the vocabulary is represented as a $K$-dimensional 
real-valued vector ${\bf w} \in \mathbb{R}^K$, as in the case of the encoder. 
Let ${\bf R}$ denote a $V \times K$ matrix of word representation vectors \footnote{Note that this is a different matrix then that used by the encoder. We use the same vocabulary throughout both models.}
where $V$ is the vocabulary size. Let $(w_1, \ldots w_{n-1})$ be a tuple of $n-1$ words where $n-1$ is the context size. The LBL model makes a linear prediction of the next word representation as
\begin{equation}
{\bf \hat{r}} = \sum_{i=1}^{n-1} {\bf C}^{(i)} {\bf w}_{i},
\end{equation}
where ${\bf C}^{(i)}, i=1,\ldots,n-1$ are $K \times K$ 
context parameter matrices. Thus, ${\bf \hat{r}}$ is the predicted representation of ${\bf w}_{n}$. 
The conditional probability $P(w_n = i | w_{1:n-1})$ of $w_n$ given $w_1,\ldots,w_{n-1}$ is
\begin{equation}
P(w_n = i | w_{1:n-1}) = \frac{\text{exp} ( {\bf \hat{r}}^T {\bf r}_i + b_i )}{\sum_{j=1}^V \text{exp} (  {\bf \hat{r}}^T {\bf r}_j + b_j )},
\end{equation}
where ${\bf b} \in \mathbb{R}^V$ is a bias vector. Learning is done with stochastic gradient descent.

\subsection{Multiplicative neural language models}

Suppose now we are given a vector ${\bf u} \in \mathbb{R}^K$ from the multimodal vector space, which has an association with a word sequence $S = \{w_1, \ldots, w_N\}$. For example, ${\bf u}$ may be the embedded representation of an image whose description is given by $S$. A multiplicative neural language model \cite{kiros2014multiplicative} models the distribution $P(w_n = i | w_{1:n-1}, {\bf u})$ of a new word $w_n$ given context from the previous words and the vector ${\bf u}$. A multiplicative model has the additional property that the word embedding matrix is instead replaced with a tensor $\bm{\mathcal{T}} \in \mathbb{R}^{V \times K \times G}$ where $G$ is the number of slices. Given ${\bf u}$, we can compute a word representation matrix as a function of ${\bf u}$ as $\bm{\mathcal{T}}^u = \sum_{i=1}^G u_i \bm{\mathcal{T}}^{(i)}$ i.e. word representations with respect to ${\bf u}$ are computed as a linear combination of slices weighted by each component $u_i$ of ${\bf u}$. Here, the number of slices $G$ is equal to $K$, the dimensionality of ${\bf u}$.

It is often unnecessary to use a fully unfactored tensor. As in e.g. \cite{memisevic2007unsupervised, krizhevsky2010factored}, we re-represent $\bm{\mathcal{T}}$ in terms of three matrices ${\bf W}^{fk} \in \mathbb{R}^{F \times K}$, 
${\bf W}^{fd} \in \mathbb{R}^{F \times G}$ and 
${\bf W}^{fv} \in \mathbb{R}^{F \times V}$, such that
\begin{eqnarray}
\bm{\mathcal{T}}^u = ({\bf W}^{fv})^{\top} \cdot \textrm{diag}({\bf W}^{fd} {\bf u}) \cdot {\bf W}^{fk}
\end{eqnarray}
where $\textrm{diag}(\cdot)$ denotes the matrix with its argument on the diagonal. These matrices are parametrized by a pre-chosen number of factors $F$. In \cite{kiros2014multiplicative}, the conditioning vector ${\bf u}$ is referred to as an \textit{attribute} and using a third-order model of words allows one to model conditional similarity: how meanings of words change as a function of the attributes they're conditioned on. 

Let ${\bf E} = ({\bf W}^{fk})^{\top} {\bf W}^{fv}$ 
denote a `folded' $K \times V$ matrix of word embeddings. 
Given the context $w_1,\ldots,w_{n-1}$,
the predicted next word representation ${\bf \hat{r}}$ is given by:
\begin{equation}
%{\bf \hat{r}} = \left( \sum_{i=1}^{n-1} {\bf C}_i ({\bf W}_{hf} {\bf W}_{f\hat{r}})^T_{w_i} \right) + {\bf C}_m {\bf x}
{\bf \hat{r}} =  \sum_{i=1}^{n-1} {\bf C}^{(i)} {\bf E}(:,w_i) ,
\end{equation}
where ${\bf E}(:,w_i)$ denotes the column of ${\bf E}$ for the word representation of $w_i$ and ${\bf C}^{(i)}, i=1,\ldots,n-1$ are $K \times K$ context matrices. Given a predicted next word representation ${\bf \hat{r}}$, the factor outputs are ${\bf f} = ({\bf W}^{fk} {\bf \hat{r}}) \bullet ({\bf W}^{fd} {\bf u})$, where $\bullet$ is a component-wise product. The conditional probability 
$P(w_n = i | w_{1:n-1}, {\bf u})$ of $w_n$ given $w_1,\ldots,w_{n-1}$ and ${\bf u}$ can be written as
\begin{equation}
P(w_n = i | w_{1:n-1}, {\bf u}) = 
\frac{\text{exp} \big( ({\bf W}^{fv}(:,i))^{\top} {\bf f} + b_i \big)}
{\sum_{j=1}^V \text{exp} \big(  ({\bf W}^{fv}(:,j))^{\top} {\bf f} + b_j \big)},
\nonumber 
\end{equation}
where ${\bf W}^{fv}(:,i)$ denotes the column of ${\bf W}^{fv}$ corresponding to word $i$. In contrast to the log-bilinear model, the matrix of word representations ${\bf R}$ from before is replaced with the factored tensor $\bm{\mathcal{T}}$ that we have derived. We compared the multiplicative model against an additive variant \cite{kirosmultimodal} and found on large datasets, such as the SBU Captioned Photo dataset \cite{ordonez2011im2text}, the multiplicative variant significantly outperforms its additive counterpart. Thus, the SC-NLM is derived from the multiplicative variant. 

\subsection{Structure-content neural language models}

\begin{figure}
%\vspace{-0.2in}
  \centering

  \mbox{
    \subfigure[Multiplicative NLM]{\includegraphics[width=0.29\columnwidth]{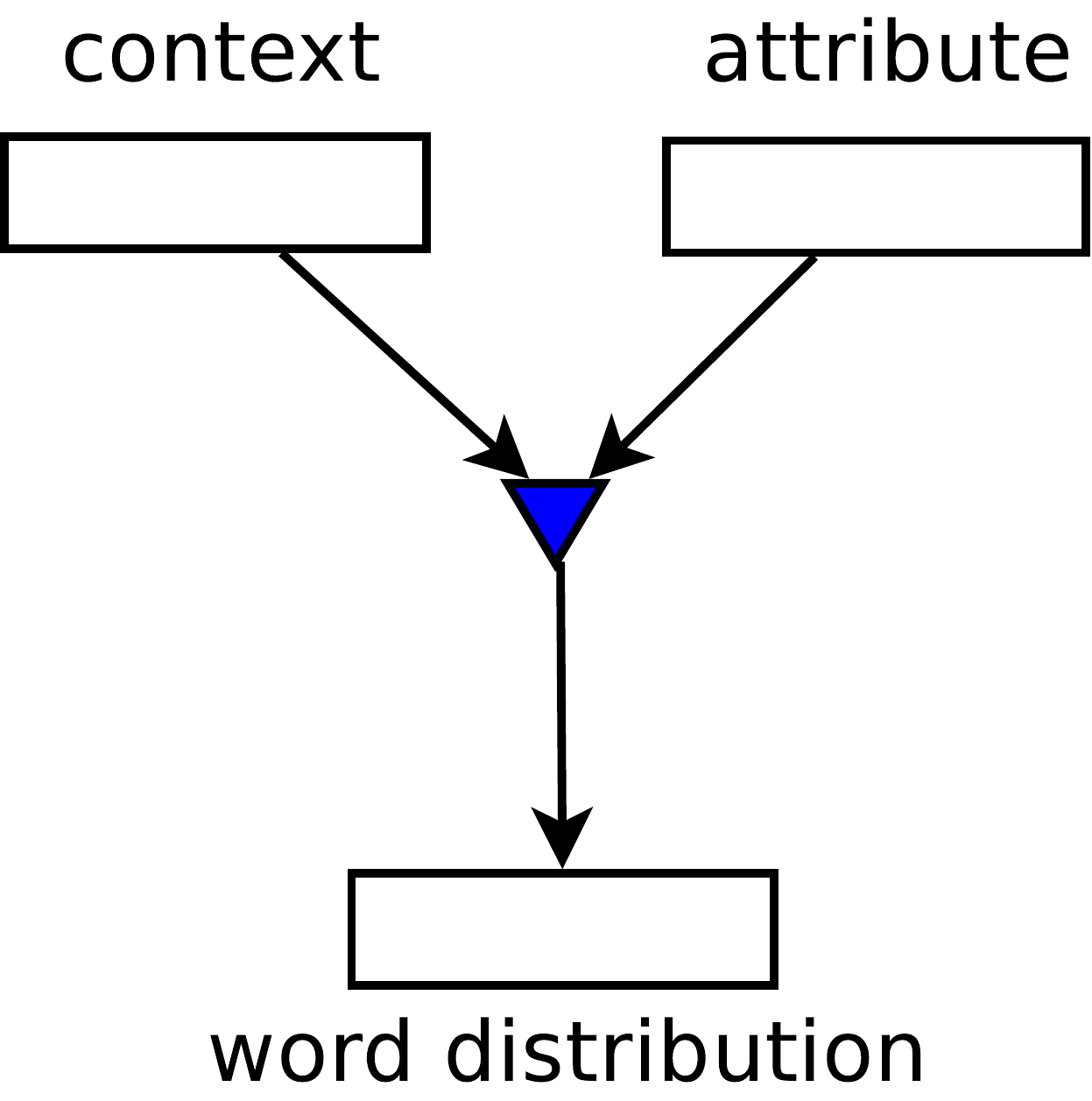}}%\quad
    \hspace{5mm}
    \subfigure[Structure-content NLM]{\includegraphics[width=0.316\columnwidth]{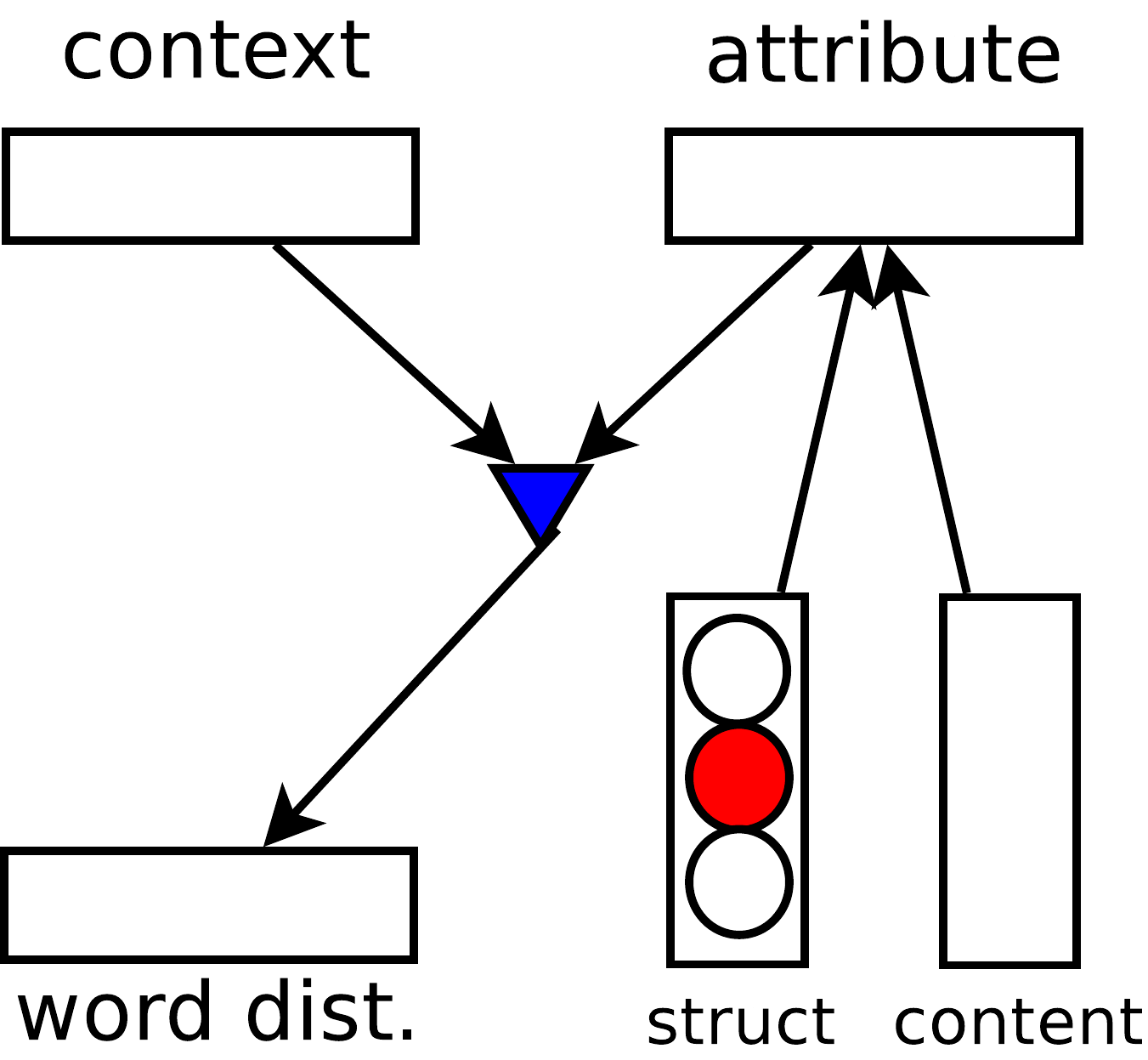}}\quad
    \hspace{2mm}
    \subfigure[SC-NLM prediction]{\includegraphics[width=0.33\columnwidth]{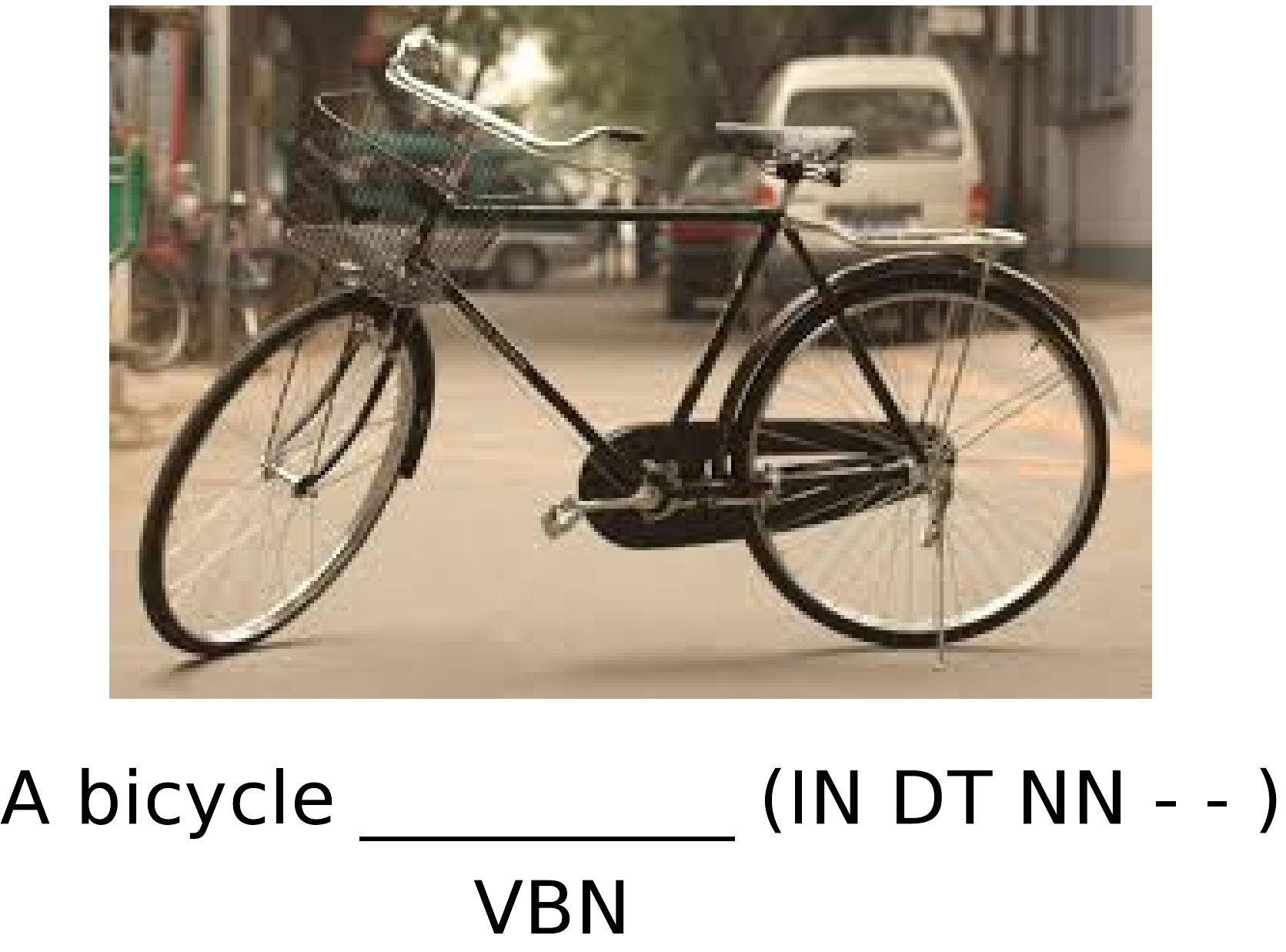}}%\quad
  }

 %\vspace{-0.1in}
  \caption{Left: multiplicative neural language model. Middle: Structure-content neural language model (SC-NLM). Right: The prediction problem of an SC-NLM.}
  \label{fig:nlms}
  %\vspace{-3mm}
\end{figure}

We now describe the structure-content neural language model. Suppose that, along with a description $S = \{w_1, \ldots, w_N\}$, we are also given a sequence of word-specific structure variables $T = \{t_1, \ldots, t_N\}$. Throughout our experiments, each $t_i$ corresponds to the part-of-speech for word $w_i$, although other possibilities can be used instead. Given an embedding ${\bf u}$ (the content vector), our goal is to model the distribution $P(w_n = i | w_{1:n-1}, t_{n:n+k}, {\bf u})$ from previous word context $w_{1:n-1}$ and forward structure context $t_{n:n+k}$, where $k$ is the forward context size. Figure $~\ref{fig:nlms}$ gives an illustration of the model and prediction problem. Intuitively, the structure variables help guide the model during the generation phrase and can be thought of as a soft template to help avoid the model from generating grammatical nonsense. Note that this model shares a resemblance with the NNJM of \cite{devlin2014fast} for machine translation, where the previous word context are predicted words in the target language, and the forward context are words in the source language.

Our model can be interpreted as a multiplicative neural language model but where the attribute vector is no longer ${\bf u}$ but instead an additive function of ${\bf u}$ and the structure variables $T$. Let $\{ {\bf t}_n, \ldots, {\bf t}_{n+k} \}, {\bf t}_i \in \mathbb{R}^K, i=n,\ldots,n+k$ be embedding vectors for the structure variables $T$. These are obtained from a learned lookup table in the same way as words are. We introduce a sequence of $G \times G$ structure context matrices ${\bf T}^{(i)}, i=n,\ldots,n+k$ which play the same role as the word context matrices ${\bf C}^{(i)}$. Let ${\bf T}_u$ denote a $G \times K$ context matrix for the multimodal vector ${\bf u}$. The attribute vector ${\bf \hat{u}}$ of combined structure and content information is computed as
\begin{equation}
{\bf \hat{u}} = \left[ \left( \sum_{i=n}^{n+k} {\bf T}^{(i)} {\bf t}_{i} \right) + {\bf T}^{(u)} {\bf u} + {\bf b} \right]_+
\end{equation}
where $[\cdot]_+ = max\{\cdot, 0\}$ is a ReLU non-linearity and ${\bf b}$ is a bias vector. The vector ${\bf \hat{u}}$ now plays the same role as the vector ${\bf u}$ for the multiplicative model previously described and the remainder of the model remains unchanged. Our experiments use $G = K = 300$ and factors $F = 100$.

The SC-NLM is trained on a large collection of image descriptions (e.g. Flickr30K). There are several choices available for representing the conditioning vectors ${\bf u}$. One choice would be to use the embedding of the corresponding image. An alternative choice, which is the approach we take, is to condition on the embedding vector for the description $S$ computed with the LSTM. The advantage of this approach is that the SC-NLM can be trained purely on text alone. This allows us to make use of large amounts of monolingual text (e.g. non image captions) to improve the quality of the language model. Since the embedding vectors of $S$ share a joint space with the image embeddings, we can also condition the SC-NLM on image embeddings (e.g. at test time, when no description is available) after the model has been trained. This is a significant advantage over a conditional language model that explicitly requires image-caption pairs for training and highlights the strength of a multimodal encoding space.

Due to space limitations, we leave the full details of our caption generation procedure to the supplementary material.

\section{Experiments}

\subsection{Image-sentence ranking}

Our main quantitative results is to establish the effectiveness of using an LSTM sentence encoder for ranking image and descriptions. We perform the same experimental procedure as done by \cite{karpathy2014deep} on the Flickr8K \cite{hodosh2013framing} and Flickr30K \cite{hodoshimage} datasets. These datasets come with 8,000 and 30,000 images respectively with each image annotated using 5 sentences by independent annotators. As with \cite{karpathy2014deep}, we did not do any explicit text preprocessing. We used two convolutional network architectures for extracting 4096 dimensional image features: the Toronto ConvNet \footnote{\url{https://github.com/TorontoDeepLearning/convnet}} as well as the 19-layer OxfordNet \cite{simonyan2014very} which finished 2nd place in the ILSVRC 2014 classification competition. Following the protocol of \cite{karpathy2014deep}, 1000 images are used for validation, 1000 for testing and the rest are used for training. Evaluation is performed using Recall@K, namely the mean number of images for which the correct caption is ranked within the top-K retrieved results (and vice-versa for sentences). We also report the median rank of the closest ground truth result from the ranked list. We compare our results to each of the following methods:

{\bf DeViSE.} The deep visual semantic embedding model \cite{fromedevise2013} was proposed as a way of performing zero-shot object recognition and was used as a baseline by \cite{karpathy2014deep}. In this model, sentences are represented as the mean of their word embeddings and the objective function optimized matches ours.

{\bf SDT-RNN.} The semantic dependency tree recursive neural network \cite{socher2013grounded} is used to learn sentence representations for embedding into a joint image-sentence space. The same objective is used.

{\bf DeFrag.} Deep fragment embeddings \cite{karpathy2014deep} were proposed as an alternative to embedding full-frame image features and take advantage of object detections from the R-CNN \cite{girshick2013rich} detector. Descriptions are represented as a bag of dependency parses. Their objective incorporates both a global and fragment objectives, for which their global objective matches ours.

{\bf m-RNN.} The multimodal recurrent neural network \cite{mao2014explain} is a recently proposed method that uses perplexity as a bridge between modalities, as first introduced by \cite{kirosmultimodal}. Unlike all other methods, the m-RNN does not use a ranking loss and instead optimizes the log-likelihood of predicting the next word in a sequence conditioned on an image.

Our LSTMs use 1 layer with 300 units and weights initialized uniformly from [-0.08, 0.08]. The margin $\alpha$ was set to $\alpha = 0.2$, which we found performed well on both datasets. Training is done using stochastic gradient descent with an initial learning rate of 1 and was exponentially decreased. We used minibatch sizes of 40 on Flickr8K and 100 on Flickr30K. No momentum was used. The same hyperparameters are used for the OxfordNet experiments.

\subsubsection{Results}

\begin{table*}[t]
\small
\centering
\begin{tabulary}{\linewidth}{L|CCCC|CCCC}
\hline
\multicolumn{9}{c}{\textbf{Flickr8K}} \\
\hline
& \multicolumn{4}{c}{Image Annotation} & \multicolumn{4}{c}{Image Search} \\
\textbf{Model} & \textbf{R@1} & \textbf{R@5} & \textbf{R@10} & \textbf{Med} \it{r} & \textbf{R@1} & \textbf{R@5} & \textbf{R@10} & \textbf{Med} \it{r} \\
\hline
\hline
Random Ranking & 0.1 & 0.6 & 1.1 & 631 & 0.1 & 0.5 & 1.0 & 500 \\
\hline
SDT-RNN \cite{socher2013grounded} & 4.5 & 18.0 & 28.6 & 32 & 6.1 & 18.5 & 29.0 & 29 \\
$\dagger$ DeViSE \cite{fromedevise2013} & 4.8 & 16.5 & 27.3 & 28 & 5.9 & 20.1 & 29.6 & 29 \\
$\dagger$ SDT-RNN \cite{socher2013grounded} & 6.0 & 22.7 & 34.0 & 23 & 6.6 & 21.6 & 31.7 & 25 \\
DeFrag \cite{karpathy2014deep} & 5.9 & 19.2 & 27.3 & 34 & 5.2 & 17.6 & 26.5 & 32 \\
$\dagger$ DeFrag \cite{karpathy2014deep} & 12.6 & 32.9 & 44.0 & 14 & 9.7 & 29.6 & 42.5 & 15 \\
m-RNN \cite{mao2014explain} & \underline{14.5} & \underline{37.2} & \underline{48.5} & \underline{11} & 11.5 & \underline{31.0} & 42.4 & 15 \\
\hline
Our model & 13.5 & 36.2 & 45.7 & 13 & 10.4 & \underline{31.0} & \underline{43.7} & \underline{14} \\
Our model (OxfordNet) & \textbf{18.0} & \textbf{40.9} & \textbf{55.0} & \textbf{8} & \textbf{12.5} & \textbf{37.0} & \textbf{51.5} & \textbf{10} \\
\hline
\end{tabulary}
\vspace{-0.1in}
\caption{{\small Flickr8K experiments. \textbf{R@K} is Recall@K (high is good). \textbf{Med} {\it r} is the median rank (low is good). Best results overall are \textbf{bold} while best results without OxfordNet features are \underline{underlined}. A $\dagger$ infront of the method indicates that object detections were used along with single frame features.}}
\label{fig:f8}
\vspace{-0.1in}
\end{table*}

\begin{table*}[t]
\small
\centering
\begin{tabulary}{\linewidth}{L|CCCC|CCCC}
\hline
\multicolumn{9}{c}{\textbf{Flickr30K}} \\
\hline
& \multicolumn{4}{c}{Image Annotation} & \multicolumn{4}{c}{Image Search} \\
\textbf{Model} & \textbf{R@1} & \textbf{R@5} & \textbf{R@10} & \textbf{Med} \it{r} & \textbf{R@1} & \textbf{R@5} & \textbf{R@10} & \textbf{Med} \it{r} \\
\hline
\hline
Random Ranking & 0.1 & 0.6 & 1.1 & 631 & 0.1 & 0.5 & 1.0 & 500 \\
\hline
$\dagger$ DeViSE \cite{fromedevise2013} & 4.5 & 18.1 & 29.2 & 26 & 6.7 & 21.9 & 32.7 & 25 \\
$\dagger$ SDT-RNN \cite{socher2013grounded} & 9.6 & 29.8 & 41.1 & 16 & 8.9 & 29.8 & 41.1 & 16 \\
$\dagger$ DeFrag \cite{karpathy2014deep} & 14.2 & 37.7 & 51.3 & 10 & 10.2 & 30.8 & 44.2 & 14 \\
$\dagger$ DeFrag + Finetune CNN \cite{karpathy2014deep} & 16.4 & \underline{40.2} & \underline{54.7} & \underline{8} & 10.3 & 31.4 & 44.5 & \underline{13} \\
m-RNN \cite{mao2014explain} & \underline{18.4} & \underline{40.2} & 50.9 & 10 & \underline{12.6} & 31.2 & 41.5 & 16 \\
\hline
Our model & 14.8 & 39.2 & 50.9 & 10 & 11.8 & \underline{34.0} & \underline{46.3} & \underline{13} \\
Our model (OxfordNet) & \textbf{23.0} & \textbf{50.7} & \textbf{62.9} & \textbf{5} & \textbf{16.8} & \textbf{42.0} & \textbf{56.5} & \textbf{8} \\
\hline
\end{tabulary}
\vspace{-0.1in}
\caption{{\small Flickr30K experiments. \textbf{R@K} is Recall@K (high is good). \textbf{Med} {\it r} is the median rank (low is good). Best results overall are \textbf{bold} while best results without OxfordNet features are \underline{underlined}. A $\dagger$ infront of the method indicates that object detections were used along with single frame features.}}
\label{fig:f30}
\vspace{-0.1in}
\end{table*}

Tables $~\ref{fig:f8}$ and $~\ref{fig:f30}$ illustrate our results on Flickr8K and Flickr30K respectively. The performance of our model is comparable to that of the m-RNN. For some metrics we outperform or match existing results while on others m-RNN outperforms our model. The m-RNN does not learn an explicit embedding between images and sentences and relies on perplexity as a means of retrieval. Methods that learn explicit embedding spaces have a significant speed advantage over perplexity-based retrieval methods, since retrieval is easily done with a single matrix multiply of stored embedding vectors from the dataset with the query vector. Thus explicit embedding methods are much better suited for scaling to large datasets. 

Perhaps more interestingly is the fact that both our method and the m-RNN outperform existing models that integrate object detections. This is contradictory to \cite{socher2013grounded}, where recurrent networks are the worst performing models. This highlights the effectiveness of LSTM cells for encoding dependencies across descriptions and learning meaningful distributed sentence representations. Integrating object detections into our framework should almost surely improve performance as well as allow for interpretable retrievals, as in the case of DeFrag.

Using image features from the OxfordNet model results in a significant performance boost across all metrics, giving new state-of-the-art numbers on these evaluation tasks.

\subsection{Multimodal linguistic regularities}

Word embeddings learned with skip-gram \cite{mikolov2013efficient} or neural language models \cite{Bengio:2003:NPL:944919.944966} were shown by \cite{mikolov2013linguistic} to exhibit linguistic regularities that allow these models to perform analogical reasoning. For instance, "man" is to "woman" as "king" is to ? can be answered by finding the closest vector to "king" - "man" + "woman". A natural question we ask is whether multimodal vector spaces exhibit the same phenomenon. Would *image of a blue car* - "blue" + "red" be near images of red cars? 

Suppose that we train an embedding model with a linear encoder, namely ${\bf v} = \sum_{i=1}^N {\bf w}_i$ for word vectors ${\bf w}_i$ and sentence vector ${\bf v}$ (where both ${\bf v}$ and the image embedding are normalized to unit length). Using our example above, let ${\bf v}_{blue}$, ${\bf v}_{red}$ and ${\bf v}_{car}$ denote the word embeddings for blue, red and car respectively. Let ${\bf I}_{bcar}$ and ${\bf I}_{rcar}$ denote embeddings of images with blue and red cars. After training a linear encoder, the model has the property that ${\bf v}_{blue} + {\bf v}_{car} \approx {\bf I}_{bcar}$ and ${\bf v}_{red} + {\bf v}_{car} \approx {\bf I}_{rcar}$. It follows that
\begin{eqnarray}
{\bf v}_{car} &\approx& {\bf I}_{bcar} - {\bf v}_{blue} \\
{\bf v}_{red} + {\bf v}_{car} &\approx& {\bf I}_{bcar} - {\bf v}_{blue} + {\bf v}_{red} \\
{\bf I}_{rcar} &\approx& {\bf I}_{bcar} - {\bf v}_{blue} + {\bf v}_{red}
\end{eqnarray}
Thus given a query image ${\bf q}$, a negative word ${\bf w}_n$ and a positive word ${\bf w}_p$ (all with unit norm), we seek an image ${\bf x^*}$ such that:
\begin{equation}
{\bf x^*} = \underset{{\bf x}}{\operatorname{argmax}} \frac{({\bf q} - {\bf w}_n + {\bf w}_p)^{\top} {\bf x} }{\| {\bf q} - {\bf w}_n + {\bf w}_p \|}
\end{equation}
The supplementary material contains qualitative evidence that the above holds for several types of regularities and images. \footnote{For this model we finetune the word representations.} In our examples, we consider retrieving the top-4 nearest images. Occasionally we observed that a poor result would be obtained within the top-4 among good results. We found a simple strategy for removing these cases is to first retrieve the top N nearest images, then re-sort these based on their distance to the mean of the N images.

It is worth noting that these kinds of regularities are not well observed with an LSTM encoder, since sentences are no longer just a sum of their words. The linear encoder is roughly equivalent to the DeViSE baselines in tables $~\ref{fig:f8}$ and $~\ref{fig:f30}$, which perform significantly worse for retrieval than an LSTM encoder. So while these regularities are interesting the learned multimodal vector space is not well apt for ranking sentences and images.

\subsection{Image caption generation}

We generated image descriptions for roughly 800 images from the SBU captioned photo dataset \cite{ordonez2011im2text}. These are the same images used to display results by the current state-of-the-art composition based approach, TreeTalk \cite{berg2014tree}. \footnote{\url{http://ilp-cky.appspot.com/generation}} Our LSTM encoder and SC-NLM decoder were trained by concatenating the Flickr30K dataset with the recently released Microsoft COCO dataset \cite{lin2014microsoft}, which combined give us over 100,000 images and over 500,000 descriptions for training. The SBU dataset contains 1 million images each with a single description and was used by \cite{berg2014tree} for training their model. While the SBU dataset is larger, the annotated descriptions are noisier and more personalized.

The generated results can be found at \url{http://www.cs.toronto.edu/\textasciitilde rkiros/lstm_scnlm.html} \footnote{These results use features from the Toronto ConvNet.}. For each image we show the original caption, the nearest neighbour sentence from the training set, the top-5 generated samples from our model and the best generated result from TreeTalk. The nearest neighbour sentence is displayed to demonstrate that our model has not simply learned to copy the training data. Our generated descriptions are arguably the nicest ones to date.

\section{Discussion}

When generating a description, it is often the case that only a small region is relevant at any given time. We are developing an attention-based model that jointly learns to align parts of captions to images and use these alignments to determine where to attend next, thus dynamically modifying the vectors used for conditioning the decoder. We also plan on experimenting with LSTM decoders as well as deep and bidirectional LSTM encoders.

\subsubsection*{Acknowledgments}

We would like to thank Nitish Srivastava for assistance with his ConvNet package as well as preparing the Oxford convolutional network. We also thank the anonymous reviewers from the NIPS 2014 deep learning workshop for their comments and suggestions.

\bibliography{nips2014}
\bibliographystyle{unsrt}

\newpage

\section{Supplementary material: Additional experimentation and details}
\label{intro}

\subsection{Multimodal linguistic regularities}

\begin{figure}[h]
%\vspace{-0.1in}
  \centering
  \mbox{
    \subfigure[Simple cases]{\includegraphics[width=0.46\textwidth]{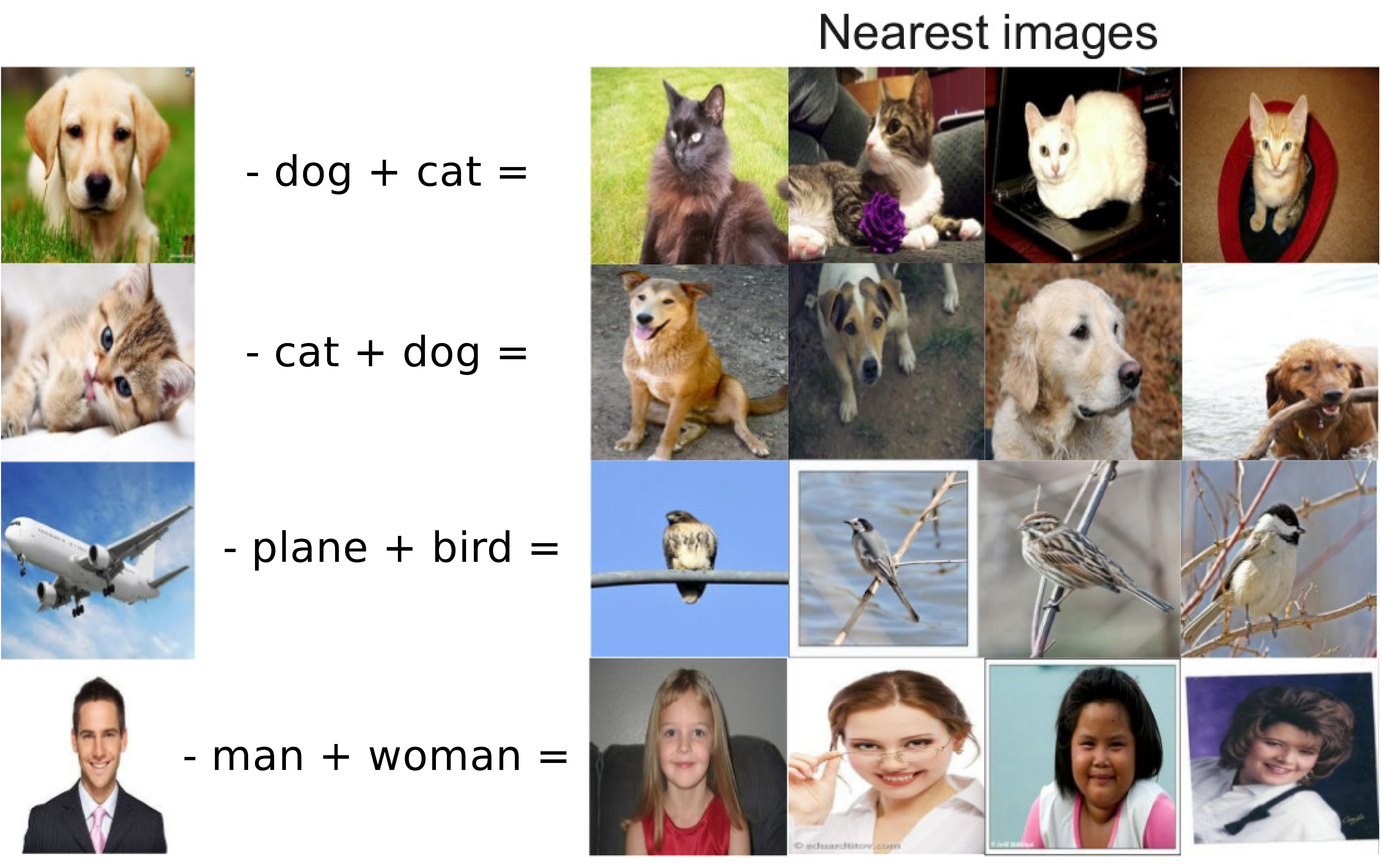}}
    \hspace{3mm}
    \subfigure[Colors]{\includegraphics[width=0.46\textwidth]{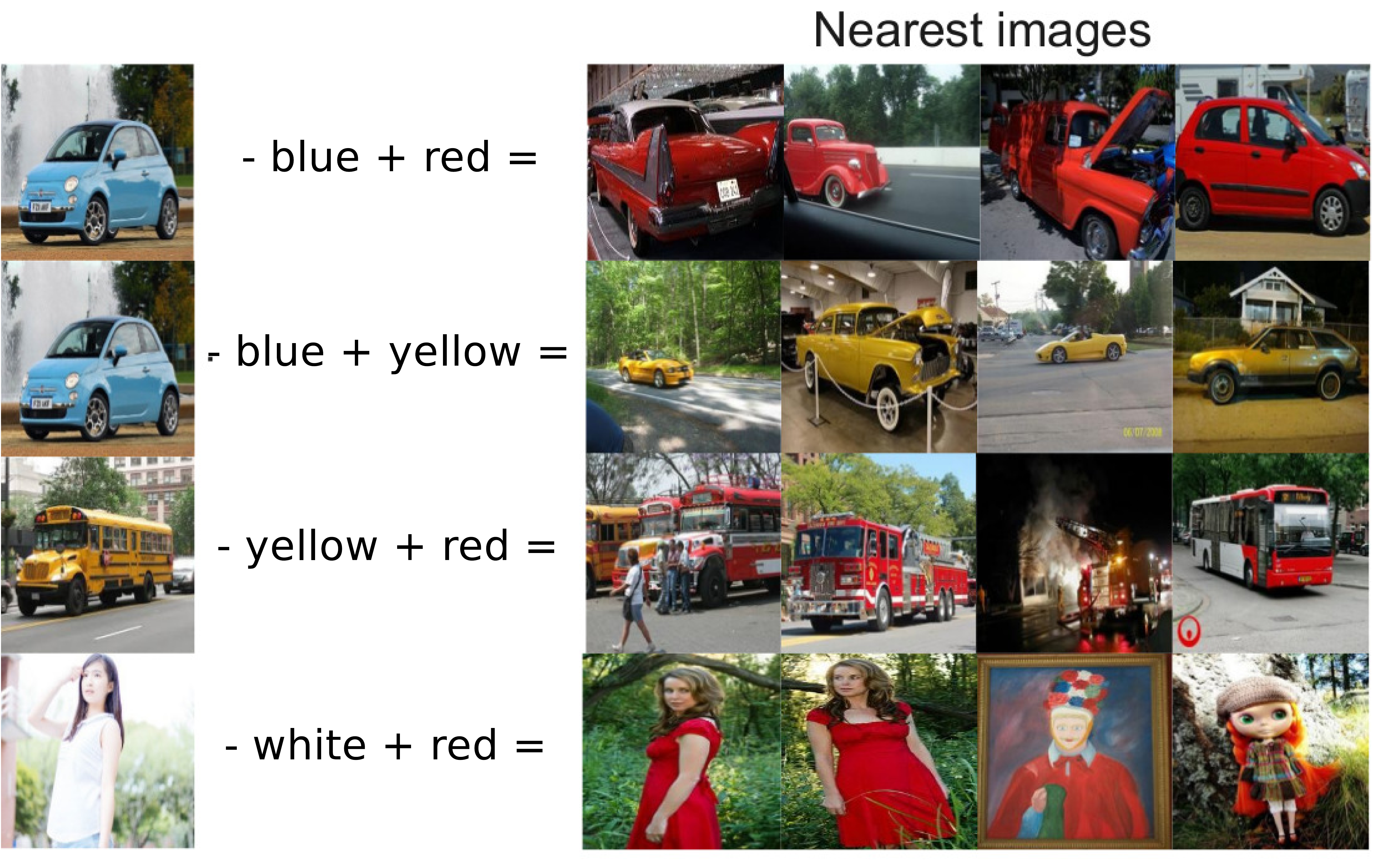}}
  }
%\vspace{-0.1in}
  \mbox{
%    \hspace{6mm}
    \subfigure[Image structure]{\includegraphics[width=0.46\textwidth]{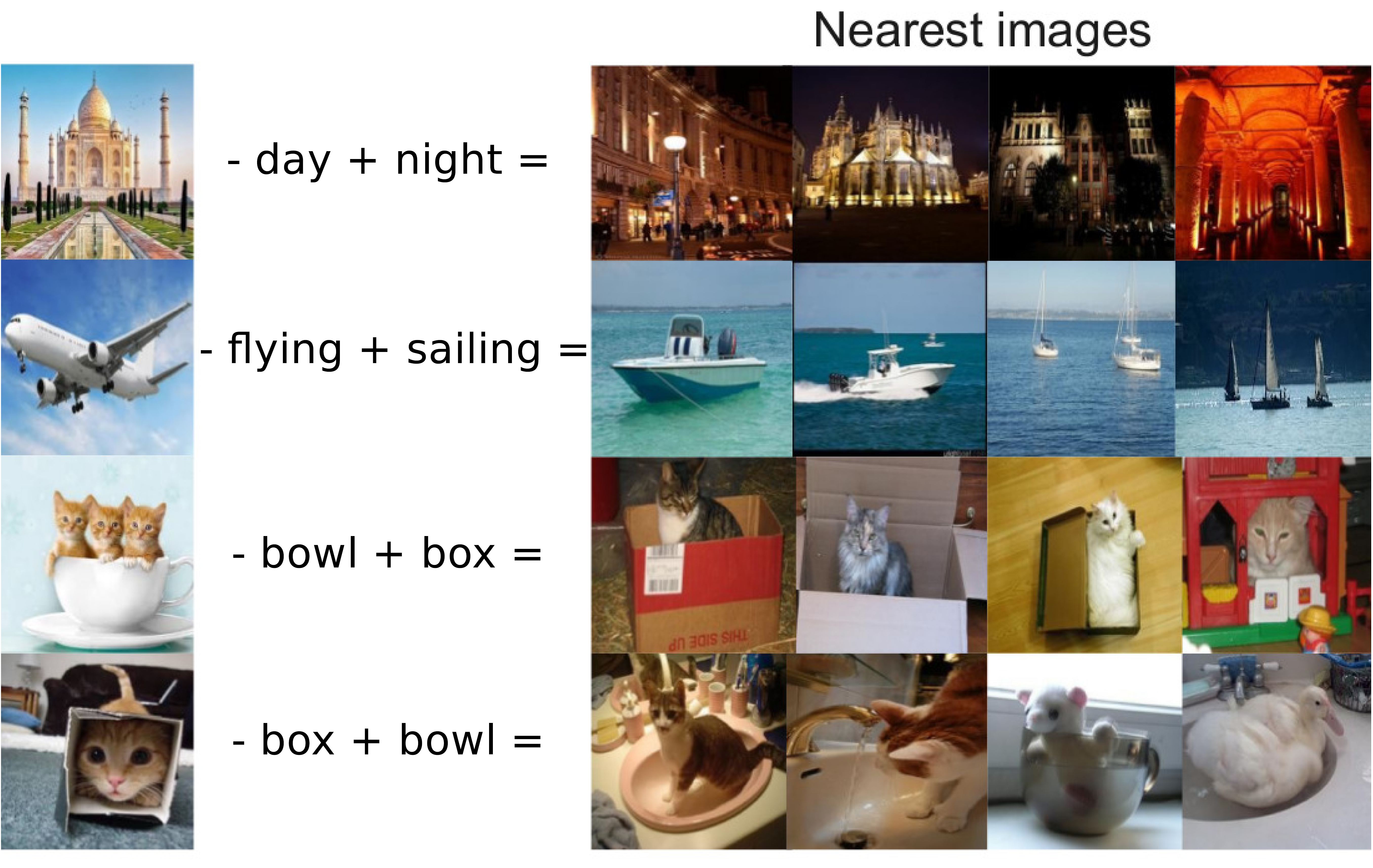}}
    \hspace{13mm}
    \subfigure[Sanity check]{\includegraphics[width=0.40\textwidth]{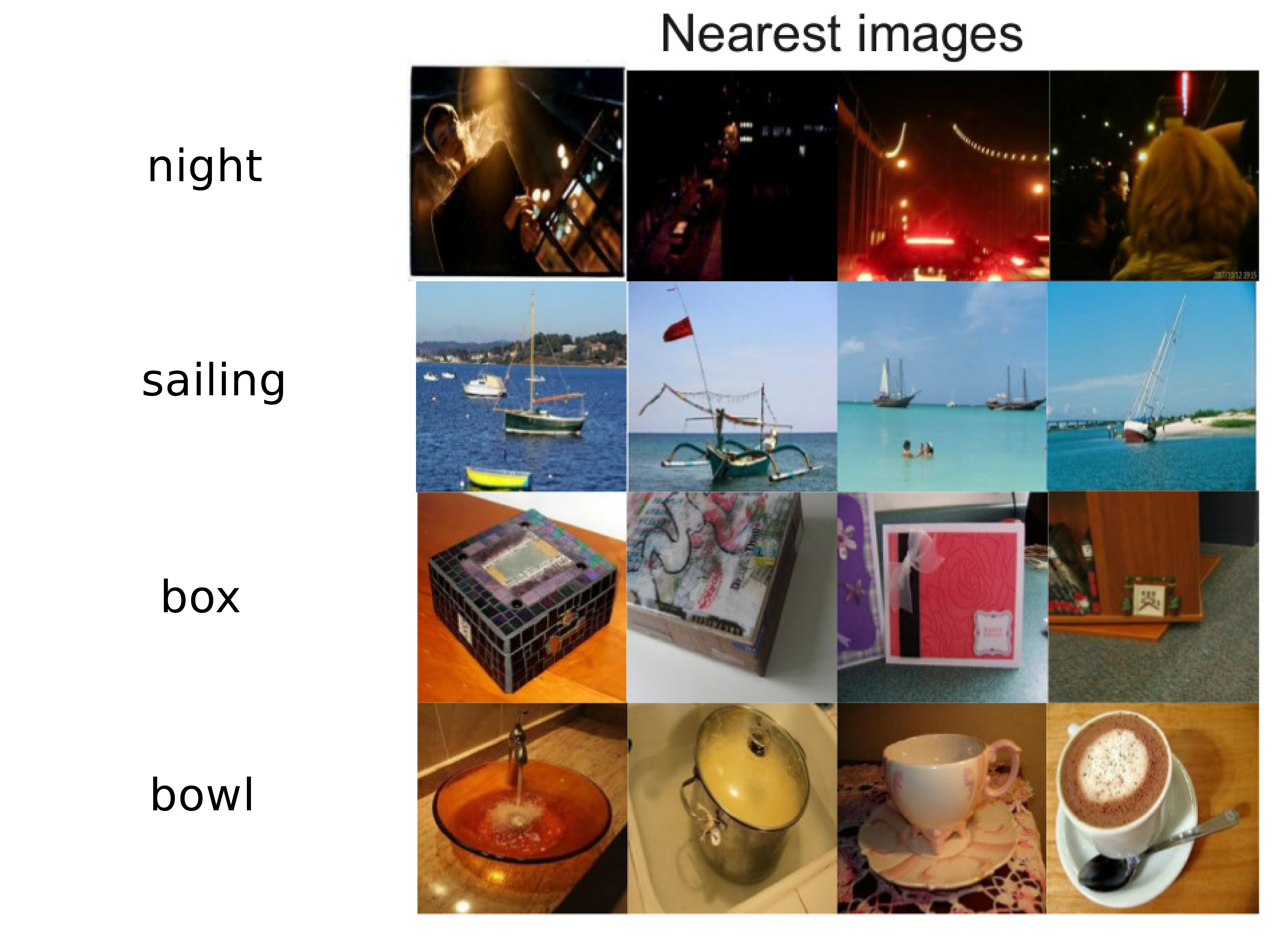}}
  }
%\vspace{-0.2in}
  \caption{Multimodal vector space arithmetic. Query images were downloaded online and retrieved images are from the SBU dataset.}
  \label{fig:lr}
%  \vspace{-3mm}
\end{figure}

\begin{figure}[h]
  \centering
  \mbox{
    \subfigure[Colors]{\includegraphics[width=0.5\textwidth]{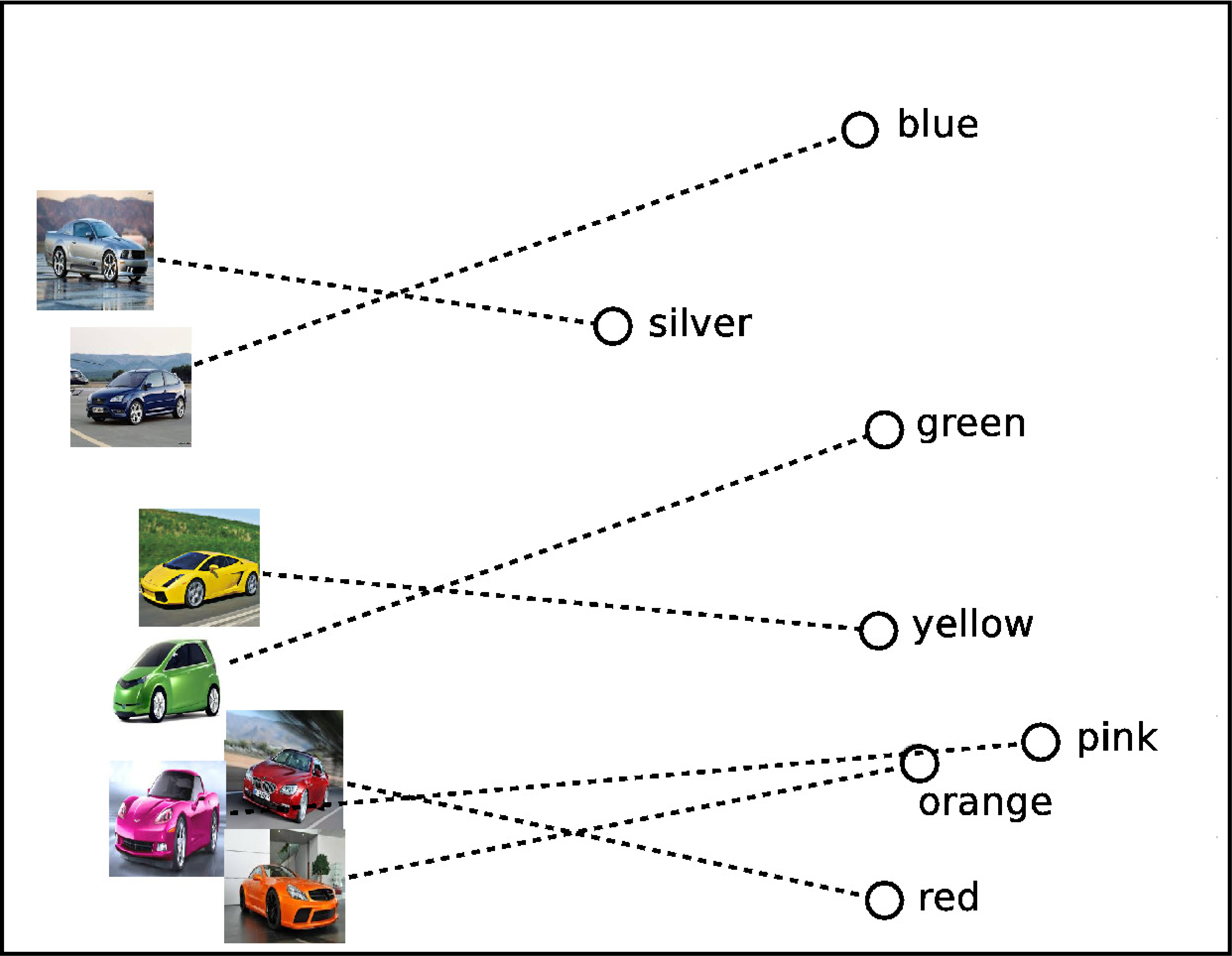}}
    \subfigure[Weather]{\includegraphics[width=0.5\textwidth]{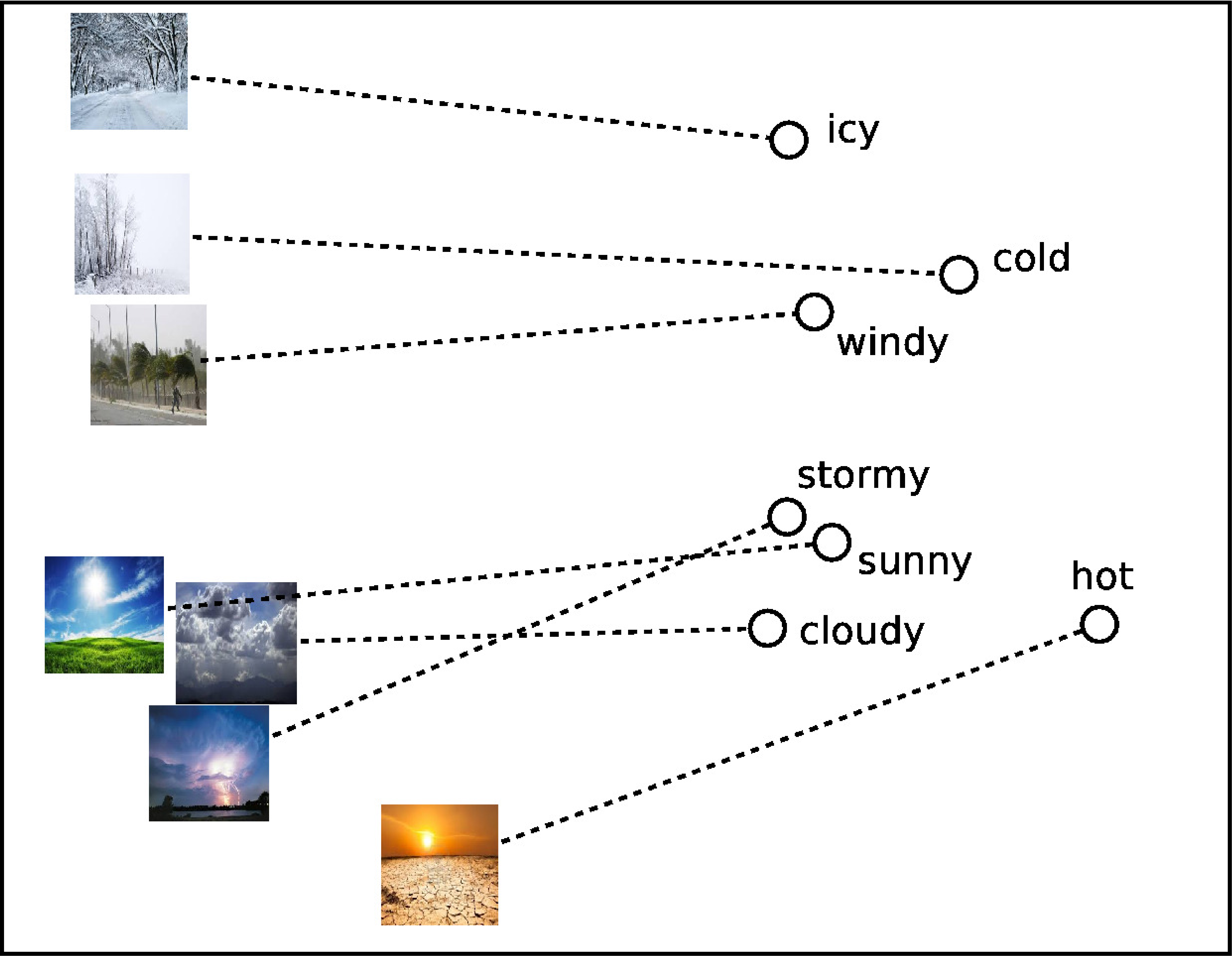}}
  }
  \caption{PCA projection of the 300-dimensional word and image representations for (a) cars and colors and (b) weather and temperature.}
  \label{fig:pca}
\end{figure}

Figure $~\ref{fig:lr}$ illustrates sample results using a model trained on the SBU dataset. All queries were downloaded online and retrieved images are from the SBU images used for training. What is of interest to note is that the resulting images depend highly on the image used for the query. For example, searching for the word `night' retrieves arbitrary images taken at night. On the other hand, an image with a building predominantly as its focus will return night images when `day' is subtracted and `night' is added. A similar phenomenon occurs with the example of cats, bowls and boxes. As additional visualizations, we computed PCA projections of cars and their corresponding colors as well as images and the weather occurrences in Figure $~\ref{fig:pca}$. These results give us strong evidence for the regularities apparent in multimodal vector spaces trained with linear encoders. Of course, sensible results are only likely to be obtained if (a) the content of the image is correctly recognized, (b) the subtraction word is relevant to the image and (c) an image exists that is sensible for the corresponding query.

\subsection{Image description generation}

The SC-NLM was trained on the concatenation of training sentences from both Flickr30K and Microsoft COCO. Given an image, we first map it into the multimodal space. From this embedding, we define 2 sets of candidate conditioning vectors to the SC-NLM:

\textbf{Image embedding.} The embedded image itself. Note that the SC-NLM was not trained with images but can be conditioned on images since the embedding space is multimodal.

\textbf{top-$N$ nearest words and sentences.} After first computing the image embedding, we obtain the top-$N$ nearest neighbour words and training sentences using cosine similarity. These retrievals are treated as a `bag of concepts' for which we compute an embedding vector as the mean of each concept. All of our results use $N = 5$.

Along with the candidate conditioning vectors, we also compute candidate POS sequences used by the SC-NLM. For this, we obtain a set of all POS sequences from the training set whose lengths were between 4 and 12, inclusive. Captions are generated by first sampling a conditioning vector, next sampling a POS sequence, then computing a MAP estimate from the SC-NLM. We generate a large list of candidate descriptions (1000 for each image in our results) and rank these candidates using a scoring function. Our scoring function consists of two feature functions:

\textbf{Translation model.} The candidate description is embedded into the multimodal space using the LSTM. We then compute a translation score as the cosine similarity between the image embedding and the embedding of the candidate description. This scores how relevant the content of the candidate is to the image. We also augment to this score a multiplicative penalty to non-stopwords that appear too frequently in the description. \footnote{For instance, given an image of a car, we would want a candidate to be ranked low if each noun in the description was `car'.}

\textbf{Language model.} We trained a Kneser-Ney trigram model on a large corpus and compute the log-probability of the candidate under the model. This scores how reasonable of an English sentence is the candidate.

The total score of a caption is then the weighted sum of the translation and language models. Due to the challenge of quantitatively evaluating generated descriptions, we tuned the weights by hand on qualitative results alone. All of the candidate descriptions are ranked by their scores, and the top-5 captions are returned.

\end{document}